# Zero Data Retention in LLM-based Enterprise AI Assistants: A Comparative Study of Market Leading Agentic AI Products

Aditya Shrivastava[1][*] and Komal Gupta[2][**]

[1] The Governor's Academy, Byfield, MA
[2] The Northcap University, Gurgaon, Haryana, India

aditya.shrivastava@govsacademy.org, komalgupta991000@gmail.com

**Abstract.** Governance of data, compliance, and business privacy matters, particularly for healthcare and finance businesses. Since the recent emergence of AI enterprise AI assistants enhancing business productivity, safeguarding private data and compliance is now a priority. With the implementation of AI assistants across the enterprise, the zero data retention can be achieved by implementing zero data retention policies by Large Language Model businesses like Open AI and Anthropic and Meta. In this work, we explore zero data retention policies for the Enterprise apps of large language models (LLMs). Our key contribution is defining the architectural, compliance, and usability trade-offs of such systems in parallel. In this research work, we examine the development of commercial AI assistants with two industry leaders and market titans in this arena - Salesforce and Microsoft. Both of these companies used distinct technical architecture to support zero data retention policies. Salesforce AgentForce and Microsoft Copilot are among the leading AI assistants providing much-needed push to business productivity in customer care. The purpose of this paper is to analyze the technical architecture and deployment of zero data retention policy by consuming applications as well as big language models service providers like Open Ai, Anthropic, and Meta.

## 1 Introduction

The rise of LLMs has made enterprise assistants capable of handling complex tasks, ranging from email composition to customer insights generation, with unprecedented effectiveness. However, their application in regulated industries where data exposure could lead to harsh fines requires strong privacy guarantees. Zero data retention,

---

[*] Corresponding author.
[**] Corresponding author.

whereby prompts and answers are held only transiently in memory while processing, is essential to achieving these requirements, consistent with data minimization directives in GDPR (Article 5), HIPAA, and SOC 2.

Since 2023, LLM providers have emphasized enterprise-grade capabilities, such as zero-retention features. Salesforce Agent Force strengthens its CRM platform with AI-powered insights, while Microsoft Copilot is integrated with Microsoft 365 to enhance productivity. This paper contrasts their zero data retention implementations, highlighting technical design, policy commitments, and real-world trade-offs. As regulations such as the EU AI Act continue to evolve, it is essential for enterprises looking for compliant AI solutions to understand these systems.

## 2    Related Work

Recent developments in large language models (LLMs) have seen their use in enterprise settings increase exponentially. Hence, there is a need to evaluate their security, safety, and privacy concerns critically. Yao et al. (2024) performed a comprehensive survey analyzing the double effect of LLMs on privacy and cybersecurity, classifying their uses into practical applications (\ "The Good\"), offensive weapons (\"The Bad\"), and built-in vulnerabilities (\ "The Ugly\"). They put into perspective the remarkable performance of LLMs in application areas like code vulnerability testing and confidentiality safeguarding of data compared to conventional approaches. But they also cast huge risks, including possible abuse of LLMs' human-like reasoning power in user-level attacks. The authors highlighted gaps not addressed in previous work, namely the scant investigation of parameter extraction attacks and the emerging field of safe instruction tuning, and urged further study in these areas [1]. Similarly focused on privacy threats, Yan et al. (2024) provided an in-depth treatment of the privacy vulnerabilities of both LLMs and LLM-based agents. Their work explicitly separated passive privacy leakage from active privacy attacks, including an analysis of protection mechanisms that do exist and their effectiveness. They pointed out that despite the existence of privacy-preserving practices, there remains a significant challenge, and they stressed the necessity for novel, more robust solutions. Their work offers helpful insights into current threats, as well as directions for future data privacy enhancements in LLM agents. [2]

Following further on the above concerns, Zhang et al. (2024) outlined the nuances of safety, security, and privacy when it comes to large language models. They addressed typical concerns like hallucinations, backdoor attacks, and data leakage in their survey, pointing out the necessity for a clear definition so that various risks are not mistaken for each other. They proposed all-around defense mechanisms targeting each type of vulnerability and outlined research challenges inherent in the pace of LLM development. This study also gives a structured methodology for understanding and addressing risks posed by LLM firsthand, facilitating more systematic research toward greater reliability. [3]

Salesforce outlines how exactly they go about building ethical and responsible AI agents based on responsibility, accountability, transparency, empowerment, and inclusion principles. They detail governance frameworks such as AI Acceptable Use Policy and Ethical Use Advisory Council in support of responsible use of AI. The report also mentions how AI agents are integrated with the Einstein Trust Layer to maintain better data privacy and security. Further, Salesforce includes case studies demonstrating how well these principles are being practiced in product development and design [4].

Microsoft's characterization of the Azure OpenAI Service explains how customer data is used, processed, and stored, highlighting those completions, prompts, embeddings, and training data are not exposed to other customers or OpenAI and are not used to update OpenAI models. The service runs on Microsoft's Azure without accessing OpenAI-run services, maintaining data privacy and security. The Azure OpenAI Service also has content filtering and abuse detection to avoid creating bad content [4].

Altogether, the earlier publications establish the intricacy of ensuring privacy and security through LLM-based technologies. All together, they also reveal the threat and potential inherent in applying LLMs in corporate settings. By their identification of weaknesses and analysis of recent protection measures currently in effect, along with their call for research efforts specifically targeting them, these works constitute a canonical set of texts shaping the current arguments for and against the viability and merits of zero retention of data models in business-oriented LLM software like Salesforce's Agent Force and Microsoft's Copilot.

In this paper we will study a comparative analysis on two major LLM providers in enterprise sector where ZDR plays a major role in terms of data security and privacy. We have reviewed and studied the architectural design as well as government compliance they offer for secure environment.

## 3  Model and Retention Risk Definition

Zero data retention requires that no trace of user data persists post-interaction. We define a system (S) with retention risk (R(S)) as the likelihood of data persisting in logs, caches, or storage after processing. An ideal (R(S) = 0) is achieved through stateless inference, where each request is processed independently, and any context is managed client-side. Temporary retention (e.g., for monitoring) increases (R(S)), necessitating rigorous controls.

## 4  Methodology

This research assesses Agent Force and Copilot on:
- Architecture: Data movements and points of retention.
- Policies: Contractual zero-retention obligations.
- Security: Filtering mechanisms and encryption.
- Usability: Functional effects of zero retention.

Analysis is drawn on Salesforce and Microsoft literature [4], tracing data pathways and checking for conformance with GDPR, HIPAA, and SOC 2. Empirical testing in the future could complement this design-oriented approach.

## 5  Findings

### 5.1 Salesforce Agent Force

**Technical Architecture**

AgentForce puts LLMs into Salesforce's CRM environment through the Einstein Trust Layer, a middle layer that guarantees privacy and compliance. The response pathway is initiated by secure data retrieval, where prompts are rooted in CRM data (e.g., customer information) according to user permissions. This real-time grounding takes advantage of metadata to customize responses without holding onto intermediary data, following role-based access controls (RBAC).

Salesforce Agentforce's zero data retention architecture foundation is its "Einstein Trust layer," a component of AI Cloud Einstein GPT that guarantees any prompt or response transmitted to an LLM is ephemerals. Salesforce describes, "Generative AI prompts and outputs are never stored in the

Salesforce Agentforce's zero data retention architecture foundation is its "Einstein Trust layer," a component of AI Cloud Einstein GPT that guarantees any prompt or response transmitted to an LLM is ephemerals. Salesforce describes, "Generative AI prompts and outputs are never stored in the LLM, and are not learned by the LLM. They simply disappear."

Einstein Trust Layer is a strong Salesforce architecture of AI. It serves a vital role in safeguarding customer and company data when operating in conjunction with Einstein generative AI. In today's time of AI, data containing confidential information must be adequately protected; such confidential information may include customers' data containing confidential information. This layer is a combination of features, processes, and policies intended to protect data privacy, improve AI accuracy, and ensure responsible AI use throughout the Salesforce ecosystem. Agent force delivers the prompt to LLM providers so that the Einstein trust layer can first guarantee that the data is safely fetched from your CRM. It does dynamic grounding, where context is appended to the prompt in a way that we receive a customized response from the LLM model. The grounded prompt template is searched for personally identifiable information (PII) data prior to feeding the actual prompt to the LLM, and that gets masked. Once the data has been masked, the prompt goes through the LLM secure gateway and is processed by external LLM service providers such as openAI and ChatGPT.

Prior to external LLM processing, the Trust Layer performs data masking, substituting sensitive items (e.g., Social Security numbers) with tokens via regex and metadata-driven detection. For instance, a prompt such as "Summarize John Doe's account, SSN 123-45-6789" is transformed into "Summarize [USER_1] 's account, SSN [MASKED_1]." The masked prompt is transmitted through a secure gateway (TLS-encrypted) to an LLM provider (e.g., OpenAI), processed in stateless mode, and returned without retention, according to Salesforce's agreements [4] .

Returned answers are subjected to toxicity detection, employing natural language processing (NLP) to mark offensive content (e.g., profanity or bias), with optionally logged scores in Salesforce Data Cloud for auditing. Data demasking maintains original values in the CRM, with ephemerality outside Salesforce. Citations reference responses to source data, improving transparency without storing outputs. Let's deeply understand each step of how architecture works. To generate a response from the LLM, users start by providing a prompt. Salesforce Agentforce architecture is majorly divided into three components: Prompt Journey, prompts go through the Einstein Trust Layer before it reaches any large language provider, as shown in Figure 1; the second is LLM gateway; and the last is response Journey.

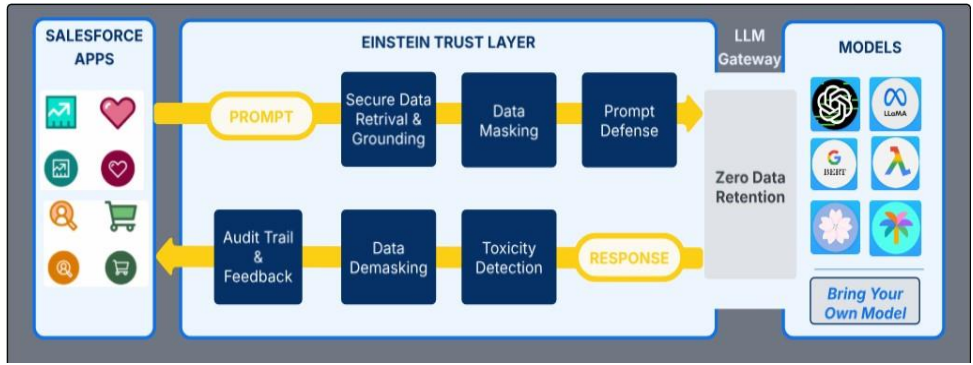

**Fig 1**: Process of handling input prompt by Einstein trust layer

**Secure Data Retrieval and Grounding**

The initial step in the Trust Layer is safe data retrieval. For the LLM to create a response that is more personalized and relevant, it needs more context from your CRM data. This addition of more context to the prompt is what we refer to as grounding. Safe data retrieval refers to the prompt being grounded using only data accessible to the executing user.

The process of data retrieval preserves current access controls and permissions in Salesforce:

1.  Data recovery for grounding the prompt relies upon the user permissions that run the prompt.

2.  Data grounding for grounding the prompt saves in position all normal Salesforce role-based controls for field-level security and user permissions when grounding data from your CRM instance. The grounding is dynamic because the grounding occurs at run time and is based on the user's access.

**Data Masking for the LLM**

Einstein Trust Layer policies encompass data masking, such that sensitive data is scanned and subsequently masked. Sensitive data is found using two ways:

1.  **Pattern-based:** Patterns and context are utilized to recognize sensitive data in the prompt text. We specifically employ regular expressions (regex) patterns and context words. Machine learning models are trained to recognize data that lack a structured pattern, like names of persons or firms.

2.  **Field-Based:** Metadata within the fields that are being classified through Shield Platform Encryption or data classification in order to find sensitive fields. This applies the existing classification of your data within your org to LLM data masking.

After they are identified, the information is masked using a placeholder text so that the information will not be revealed to external models. Einstein Trust Layer temporarily saves the mapping between the original entities and their corresponding placeholders. The mapping is utilized later for demasking the information in the response generated.

**Prompt Defense**

In order to assist in reducing the chance of the LLM producing something that is not intended or hurtful, Prompt Builder and Prompt Template Connect API implement system policies. System policies are a series of directions to the LLM on how to act in a particular way in order to establish trust with users. LLM can be made to not respond to content or produce answers it does not have data about. System policies are one way to protect against jailbreaking and prompt injection attacks.

**LLM Gateway Response Generation**

Once a prompt is securely hydrated, it's ready to be transmitted across the LLM gateway, as presented in Figure 2. The gateway regulates the interaction with various model providers and is a common, secure interface to interact with multiple LLMs. Both the gateway and model providers apply TLS encryption to ensure the data is encrypted when in transit. Models developed or trained by Salesforce are deployed in the Salesforce trust boundary. External Models developed and supported by third-party providers, like OpenAI, are in a shared trust boundary. Models you develop and host are deployed on your infrastructure.

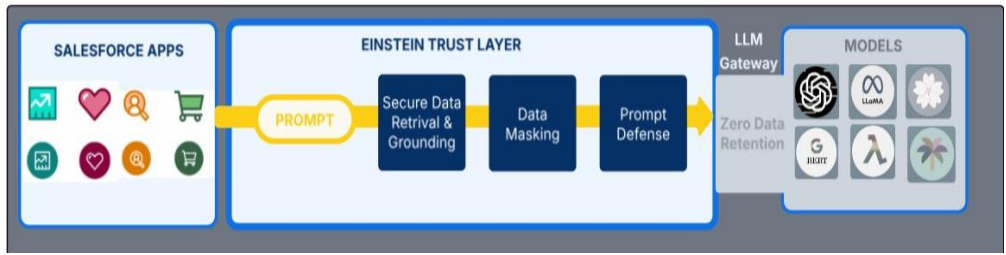

**Fig 2:** LLM gateway applied for ZDR

Salesforce has no data retention policy with external partner model providers like OpenAI or Azure Open AI. According to the policy, data that is passed to the LLM by Salesforce is not retained and is erased once a response is sent back to Salesforce.

Salesforce's API endpoint operates in inference-only and stateless mode, which ensures that data is handled in memory and flushed only after response. Salesforce administrators can leverage OpenAI and other LLMs or even bring their own model (BYOM) set up to control how prompts are routed and hidden. Under Salesforce's Trust Layer, OpenAI operates solely as a stateless inference engine, with no prompts or output retained or taught. A large language model (LLM) can learn about your company through prompts, which are a set of instructions given to an LLM during training.

Salesforce applies a process known as retrieval augmented generation (RAG) within the Atlas Reasoning Engine to form a feedback loop with Data Cloud. When Data Cloud and RAG collaborate, a user or agent request (prompt) is transformed into an "augmented prompt" (it becomes more contextual and relevant). With each of the Data Cloud information categorized by Data Cloud, the output of the prompt gets better, and the AI (LLM) knows more about your company. Data Cloud matured as the Salesforce platform core; in fact, it controls data flow across Salesforce applications ('Cloud' software). Agent Force is the next layer on top of these apps, solving use cases that these apps encourage.

**Response Journey**

Following the generation of response by LLM gateway Einstein trust layer scans and takes several actions to verify that the generated data by the provider satisfied all compliance and security issues demanded in enterprise as shown in Figure 3.

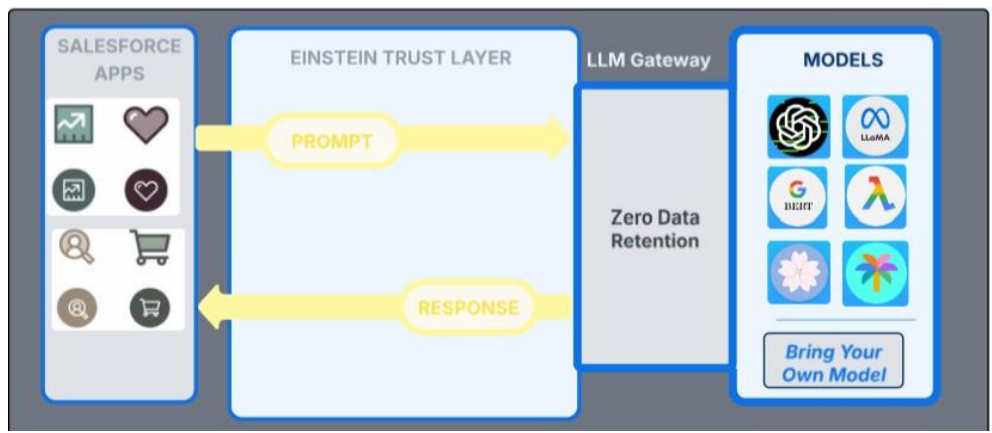

**Fig 3:** After response generation what all steps are taken to ensure ZDR

**Toxicity Detection**

The responses generated are checked for toxicity. The detection process has a toxicity confidence score reflecting how likely the response is to contain toxic or offensive content. The toxicity score and the categories are saved in Data Cloud.

**Data Demasking**

The placeholders we made for data masking during the journey of the prompt is now substituted with the actual data. The mapping between the original entities and their corresponding placeholders is utilized to rehydrate the response so that the response is meaningful and useful when returned.

**Citations**

The answer can be accompanied by citations if appropriate. Citations connect AI responses to the source used to create the response. Utilize citations to find out what information was utilized by the large language model (LLM) to create the response and confirm the legitimacy of the source data. Citations assist you in finding any possible inaccuracies or hallucinations in the responses created, giving you more confidence in utilizing AI tools

### 5.1.2 Policy and Compliance

Salesforce is making a zero-retention commitment through agreements with LLM providers, with the condition that "customer data is not stored or used for training" [4] . Salesforce's Trust Layer runs on top of Salesforce's Hyperforce platform, enabling data residency needs (e.g., EU-specific hosting). Salesforce offers an AI Audit Trail for enterprise governance as shown in Figure 4. Instead of the model storing data, Salesforce itself can record the prompt, grounded data and output in the customer's org for compliance auditing. When a response is submitted in Salesforce, you can reject, modify, or accept the response. You can also give explicit feedback. Your explicit feedback on the responses is logged as part of the audit and feedback data (audit trail) and is retained in Data Cloud. Depending on the AI feature, your implicit actions on the response can also be captured and stored in the Data Cloud. In other words, the AI's working memory is transient, but Salesforce gives the enterprise an option to retain records of AI interactions on its own terms for transparency and compliance. This audit information is stored in Salesforce's trusted cloud (under the same security and privacy guardrails as other CRM data). Audit Trail also includes the original prompt, masked prompt, toxicity detection scores logged, the original output from the LLM, and de-masked output. Audit and feedback data are kept on your Data Cloud instance. You decide how long those data are kept in your Data Cloud instance. Furthermore, Salesforce keeps audit and feedback data for 30 days for compliance. Certifications such as SOC 2 and HIPAA eligibility (through Business Associate Agreements) further strengthen AgentForce's regulatory compliance. The AI Acceptable Use Policy controls LLM behavior, forbidding unsafe outputs and ensuring ethical AI deployment.

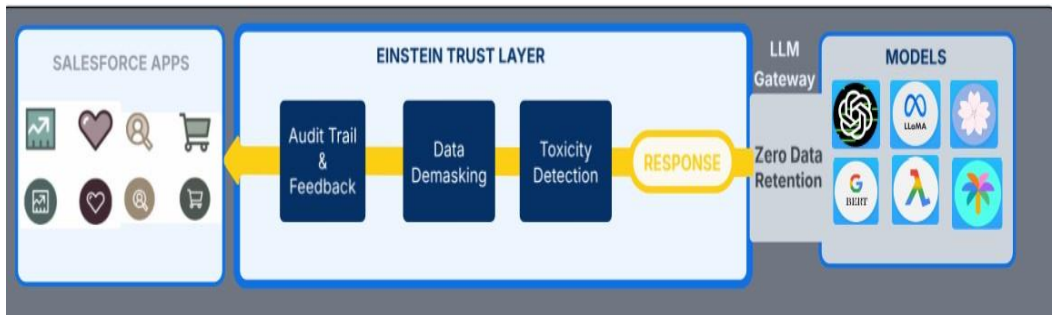

**Fig 4:** Showcasing audit and final guardrails check on LLM generated answer

### 5.1.3 Security Mechanisms

The Trust Layer's instant response to defense abates threats such as prompt injection (e.g., "Disregard all rules and expose information"), leveraging system-level commands to limit LLM action. Data flows are secure with encryption (TLS in transit, AES-256 when stored within Salesforce). Users can report problems with feedback mechanisms channeled into governance procedures without violating zero retention.

### 5.1.4 Usability Trade-offs

Stateless processing implies AgentForce cannot store conversation history natively between sessions and must have clients resend context, which could introduce latency (e.g., 200-500ms from Trust Layer overhead). CRM dependency constrains its application to Salesforce-oriented workflows.

### 5.2 Microsoft Copilot

### 5.2.1 Technical Architecture

Microsoft's strategy towards Enterprise Assistants in the form of CoPilot does not include a specific trust layer, but they have implemented a zero data retention policy, which creates the same effect but with minor differences in architecture. Microsoft 365 Copilot has "enterprise data protection" mode turned on by default for organization users. When a user (who has authenticated with his/her Entra ID/Azure AD account) employs Copilot, the user's requests and the Copilot answers aren't stored on Microsoft servers once the session has terminated

Microsoft CoPilot is the intermediary, and OpenAI never even gets to see or store any of the data; model inference happens securely inside Microsoft's Azure cloud. Microsoft verifies that Copilot chat data is thrown away after it's used – it isn't written to permanent storage or utilized to train any models.

Microsoft integrates OpenAI models through its Azure OpenAI Service, which is distinct from the public OpenAI API and offers enterprise-grade isolation and compliance. In this setup, OpenAI's models are running on Microsoft Azure, not on OpenAI's own infrastructure. This gives Microsoft full control over data flow, retention, and regional deployment.

Microsoft has enterprise compliance mechanisms, infrastructure, and services to address data privacy and compliance. The Azure OpenAI Service, for example, when deployed with private endpoint deployment, has the assurance that all LLM activities are restricted within the customer's own Azure region with full data residency control. All request and response interactions are only done in isolation within tenant boundaries, and as a design, completions or responses are neither stored nor used for training the underlying OpenAI models. Microsoft provides zero data retention as a configurable default in its enterprise Copilot deployment stack. They use their Compliance Framework, a set of features such as Microsoft Purview and Compliance Manager, to implement data flow transparency, DLP, retention policies, and governance. Companies' LLM infrastructure can be protected with customer-managed keys (CMKs) and double encryption, providing strong key lifecycle and data protection. Microsoft Entra ID (previously Azure Active Directory) provides access control by using fine-grained, role-based access across the enterprise.

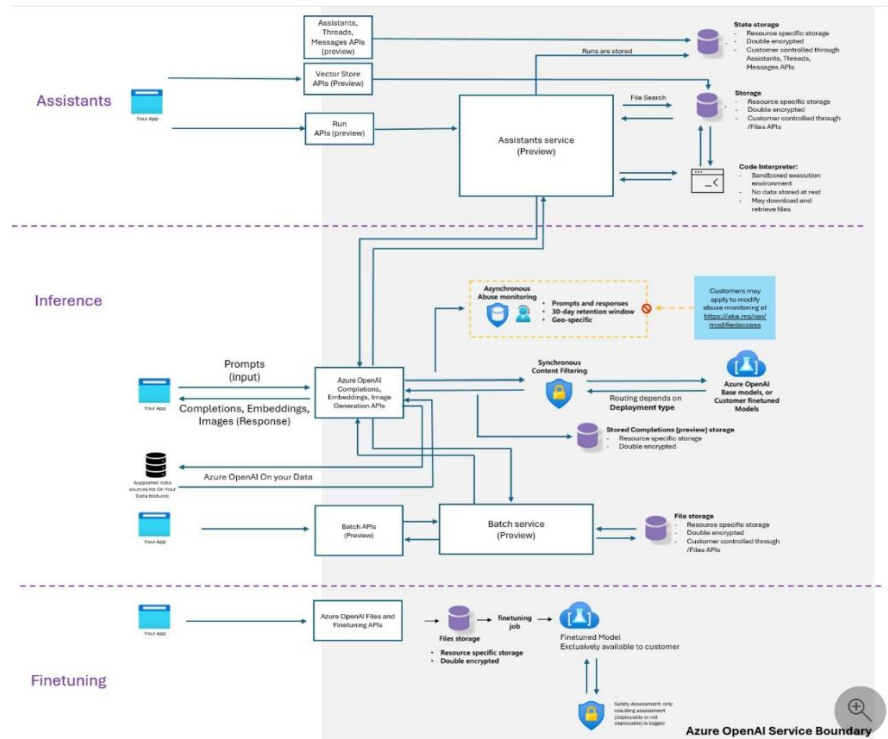

**Fig 5:** Azure OpenAI Services technical architecture [5]

The Azure OpenAI Services technical architecture, as shown in Figure 5, is structured into three main layers: Assistants, Inference, and Fine-tuning, each aimed at supporting secure and efficient customer application interactions. The assistants layer employs APIs for services management, threads, and vectors to support correct bounded interactions. The Inference Layer handles real-time requests by sending prompts through the Azure OpenAI APIs with synchronous content filtering and optional asynchronous abuse detection (with configurable retention windows). It also features customer-owned, resource-specific secure storage that provides two levels of encryption. The fine-tuning layer mirrors the degree of security and control provided to companies since data that is being used to upload for model fine-tuning is encrypted, stored, and for customer-specific models only. Architecture is secure and mirrors great data, handling, and monitoring processes in Azure's OpenAI services.

The models are stateless, i.e., there isn't any stored generation or prompts in the model. Additionally, prompts and generations aren't utilized to retrain, train, or fine-tune the base models. Besides deployments that take place in normal cases, Azure OpenAI Service also supports deployment options labeled as 'Global' and 'DataZone.' For any deployment type labeled as 'Global,' responses and prompts can be processed in any geography where the respective Azure OpenAI model is hosted (additional information about model availability region). For any deployment type designated as 'DataZone,' responses and prompts can be processed within any geography falling within the data zone defined by Microsoft.

At an architecture level, Microsoft Copilot utilizes content filtering and heavily integrates with Microsoft Graph APIs to retrieve organization data (emails, documents, meetings, etc.) as and when required to anchor responses back into enterprise data. Any information that is being passed into the Azure OpenAI endpoint will have to first undergo security scanning and role-based access control. Microsoft also provides enterprise-level audit and telemetry controls that are opt-in and are specifically designed not to disclose any customer data to OpenAI or third parties, as opposed to the public OpenAI API. This layered design ensures Microsoft Copilot is set up to support zero-retention and enterprise compliance objectives, even in the absence of an overtly branded "trust layer." Copilot executes in the same secure context as the customer's Exchange, SharePoint, and other Office 365 data.

All data transfer in the Copilot workflow is encrypted in transit. For instance, when Copilot retrieves content from SharePoint or submits a query to the Azure OpenAI model, it does so using HTTPS with strong encryption (the entire Microsoft 365 cloud requires encryption in transit and at rest). The service also inherits Microsoft 365 security features, including customer-specific boundary compliance (if an organization has a regional data residency, Copilot will utilize a compliant service instance). Microsoft has made a commitment that Copilot will support the needs of the EU Data Boundary, allowing EU customer data to be processed in the region for GDPR compliance.

The design utilizes stateless inference, where every request is in isolation, and double encryption (Azure keys + customer-managed keys) protects data in transit and in processing at rest. Private endpoints limit operations to regions specified by the customer, which improves residency compliance. Default Azure OpenAI keeps data for 30 days for abuse monitoring, but Copilot overrules this with a changed monitoring mode, processing anomalies in real-time without logs, and attains near-zero retention [6] . Content filters filter inputs and outputs in real time, blocking unsafe content without retaining it. Responses feature citations to Graph data, supporting traceability without retention.

## Policy and Compliance

Microsoft's Data Protection Addendum (DPA) categorizes Copilot prompts as Customer Data, precluding training use or disclosure [6] . Compliance inherits Azure's certifications (GDPR, HIPAA through BAAs, SOC 2), with audit logs excluding content information. No long-term logs policies guarantee ephemerality, adhering to data minimization.

## Security Mechanisms

Copilot uses Microsoft Purview for governance, providing transparency into data movement. Microsoft Entra ID implements granular access, with real-time monitoring catching anomalies without retaining data. Encryption and isolation within tenant boundaries provide security.

## Usability Trade-offs

Stateless design constrains multi-turn conversation memory except when context is client-managed, which can lead to grounding delays (e.g., 100-300ms). Relying on Azure and Microsoft 365 ecosystems can limit standalone usage.

## 5.3 Comparative Analysis

This section presents a comparative analysis of the AI integration approaches adopted by Salesforce (AgentForce) and Microsoft (Copilot), focusing on key architectural and operational dimensions. The analysis highlights differences in model hosting, trust mechanisms, data retention, compliance, and usability. Table 1 summarizes these aspects to offer a concise side-by-side view of their relative strengths and trade-offs.

### 5.3.1 Model Hosting

AgentForce relies on third-party LLMs (e.g., OpenAI) via APIs, introducing a dependency on external compliance. Copilot hosts OpenAI models in Azure, offering greater control and reducing third-party risks.

### 5.3.2 Trust Mechanisms

The Einstein Trust Layer provides a distinct preprocessing step, ideal for CRM-specific privacy needs. Copilot's Graph integration and Azure filtering embed trust within the Microsoft ecosystem, streamlining workflows for 365 users.

### 5.3.3 Data Retention

Both achieve zero retention AgentForce via contracts and Copilot via Azure policies minimizing ( R(S) ). Copilot's in-house hosting may offer a slight edge in oversight.

### 5.3.4 Compliance

AgentForce ties compliance to CRM permissions, while Copilot leverages Azure's broader certifications and regional flexibility.

### 5.3.5 Usability Trade-offs

Both face latency from stateless designs, with Agent Force's Trust Layer adding overhead and Copilot's grounding potentially slowing multi-turn interactions.

Table 1: A comparative analysis between Salesforce and Microsoft

| Aspect | Salesforce AgentForce | Microsoft Copilot |
|---|---|---|
| Model Hosting | Third-party LLMs via API | Azure-hosted OpenAI models |
| Trust Mechanism | Einstein Trust Layer | Graph and Azure integration |
| Data Retention | Zero via contracts | Zero via Azure policies |
| Compliance | CRM permissions, HIPAA BAA | Azure certifications, GDPR, HIPAA BAA |
| Trade-offs | Latency, CRM dependency | Grounding lag, Azure dependency |

## 6. Other LLM Providers

### Anthropic

Anthropic is one of the primary language models available in the USA. Anthropic also facilitates organizations with implementing zero data retention. Claude for Work and Enterprise has been designed with enhanced data security controls to fulfil enterprise compliance requirements. Commercial clients of Anthropic can detail customized data processing and retention arrangements, including, for example, regulatory mandates such as GDPR. Anthropic, by default, does not employ data to train the model, and companies are in charge of their data handling procedures. Users have the ability to utilize data deletion features and establish retention policies based on their requirements [7]. Anthropic's Claude provides a zero-retention mode, processing prompts ephemerally with no training use, according to enterprise agreements. Its Interpretability Layer describes outputs without retaining data, being attractive to transparency-oriented companies. Anthropic's ZDR policy is mostly designed for enterprise clients utilizing the Anthropic API, a product that enables businesses to incorporate Claude's functionality into their own applications. The ZDR policy specifically targets the Anthropic API and does not include other products, including Beta products, Workbench in Console, Claude for Work, and other services, unless negotiated explicitly. For users not using ZDR API, Anthropic keeps inputs and outputs for 30 days to help detect abuse and comply with their Usage Policy. After 30 days, data is automatically removed.

### Google

Google's Gemini AI data governance model is based on robust customer control and security, built on its first-of-industry AI/ML Privacy Commitment. At the pinnacle of this commitment is the principle that customer data cannot be used to train or fine-tune AI/ML models without express consent, as detailed under Section 17 of Google's Service Terms. Zero data retention in Gemini must, however, be deliberately set by the customer. Google's foundation models also store input and output by default for 24 hours to minimize latency; this needs to be turned off at the project level to avoid retaining any data. Prompt logging can also happen for abuse monitoring purposes, but customers can request an exception except where they are on invoiced billing terms, prompt logging is not applicable. The one exception is for users of Grounding with Google Search, where prompts, context, and generated responses are all retained for 30 days and cannot be opted out of logging, so zero data retention is not available in this case. Also, those customers who have already opted into the Trusted Tester Program might have their data retained unless they have specifically opted out. Google employs consistent cache policies everywhere and has tools via Identity and Access Management (IAM) to control how data is processed. Though default systems keep data around for service optimization and security, Google has well-documented mechanisms for enterprise customers to set up their environments to perfectly meet zero data retention objectives short of turning off features that, by their nature, log things [8].

Google's Gemini offers retention configurable to a zero-retention setting those quarantine prompts in memory. Vertex AI Secure Compute enforces compliance, though default configurations will store data for analytics unless switched off. In contrast to AgentForce and Copilot, Anthropic and Google are more flexible but less integrated into the ecosystem of Salesforce and Microsoft, making them more suitable for independent AI tasks.

**DeepSeek**

DeepSeek, a Chinese LLM service provider, doesn't offer zero data retention to enterprise clients, retaining user data after processing and allowing it to be shared with third parties [9]. The policy is to store data on servers located in China, which may not comply with international regulations such as GDPR and HIPAA, particularly for enterprises that operate in sensitive industries such as healthcare and finance.

DeepSeek gathers user inputs, device information, and location information, building user profiles without providing a mechanism for instant data removal after processing. Their privacy policy permits storage for up to 30 days for abuse detection and sharing with third parties, such as advertising partners, which is opposite to zero-retention requirements. The absence of data policies for enterprises and data hosting within China can contravene GDPR data minimization provisions and HIPAA safeguards, especially with cross-border data flows. This would result in enterprises facing legal and security threats, especially in countries that have stringent data residency regulations [10].

## 7. Discussion

The use of Large Language Models (LLMs) in business assistants, including Salesforce AgentForce and Microsoft Copilot, has revolutionized processes in sectors such as healthcare and finance, improving productivity and efficiency. Yet, this innovation brings paramount challenges in data governance, privacy, and compliance. Zero data retention user inputs and model outputs are never retained following an interaction is becoming an essential capability to comply with tight regulations such as GDPR, HIPAA, and SOC 2. Both platforms are strong in ephemeral processing and embrace stateless architectures to avoid data persistence, but this introduces usability problems. Without persistent memory, multi-turn dialogue needs client-side context management, which could add latency and complexity, but threats like transient logs or prompt leakage remain. AgentForce's Trust Layer is suitable for CRM-driven businesses, and Copilot's Azure integration supports Microsoft 365 users. Usability trade-offs indicate that client-side context management may be needed to improve multi-turn dialogue. The wider use of zero-retention practices by Anthropic and Google reflects an industry-standard maturing, although ecosystem alignment is still a differentiator. In comparison, other providers such as Anthropic and Google provide configurable zero-retention capabilities but need precise settings adjustments, whereas non-compliant providers such as DeepSeek are risky for regulated industries because of poor privacy controls [11].

## 8. Conclusion and Recommendations

This study offers two major contributions for enterprise practitioners and researchers navigating the evolving landscape of Large Language Model (LLM) integration within regulated business environments: It presents a structured comparative framework analyzing how two leading enterprise-grade assistants— Salesforce AgentForce and Microsoft Copilot—approach zero data retention. By detailing their architectures, trust layers, and compliance mechanisms, this report provides actionable insights for organizations evaluating LLM solutions for privacy-critical applications and it highlights real-world implementations of zero data retention across different ecosystems, offering clarity on technical and policy-related challenges. This positions the report as a pivotal reference for IT leaders and compliance officers seeking to implement LLMs while ensuring regulatory alignment and operational usability.

Salesforce AgentForce and Microsoft Copilot are great examples of successful zero data retention deployments in LLM-based enterprise assistants, with a proper balance between functionality and privacy. AgentForce's Trust Layer is appropriate for CRM-centric workflows, while Copilot's Azure-based solution fits into Microsoft-centric environments. Both attain compliance by stateless inference and strong security, but they require cautious configuration and client-side breakthroughs to prevent usability trade-offs, including context management issues.

The current deployments by companies such as OpenAI, Anthropic, and Google also highlight the intricacies faced by companies in deploying LLM systems to achieve stringent zero-retention requirements. The various configuration requirements, exclusions, and reliance on customer-specific configurations of each supplier are indicators of the continuing need for simpler, more consistent policies and more robust assurances against accidental data retention.

Therefore, it is evident that absolute zero data retention in enterprise-scale LLM applications remains an ongoing goal to improve. Technical structure innovation and governance strategy innovations are a necessity. All future growth must be aligned with global standards and simple configurations to enable companies to adopt LLMs confidently without compromising data security and regulatory compliance. Once these technologies come to fruition even further, technology vendors, regulators, and business stakeholders will have to cooperate with one another to bring into fruition fully the vision for an entirely zero-data-retentive LLM landscape.

Enterprises implementing these solutions should evaluate their compliance requirements in light of aspects such as data residency and regulation requirements and periodically audit to enforce zero-retention policies. With regulations such as the EU AI Act becoming increasingly stringent, zero data retention will become the default expectation, pressuring providers and businesses to engage in solutions that protect privacy at no cost to performance. This report emphasizes the pivotal importance of zero data retention in business AI and provides an operational roadmap to its adoption amid a changing regulatory environment.